\documentclass[final,5p,times,twocolumn]{elsarticle}
\usepackage{amssymb}
\usepackage{amsmath}
\usepackage{booktabs}
\usepackage[table,dvipsnames]{xcolor}
\usepackage{multirow}
\usepackage{makecell}
\usepackage[hidelinks]{hyperref} 
\newcommand{\hlg}[1]{\cellcolor[rgb]{0.90,0.90,0.90}#1} 

\definecolor{winner}{RGB}{31,119,180}
\definecolor{runner-up}{RGB}{255, 126, 15}

\usepackage{xcolor}

\journal{Pattern Recognition Letters}

\begin{document}

\begin{frontmatter}



\title{Emergent Morphing Attack Detection in Open Multi-modal Large Language Models\tnoteref{t1}} 


\author{Marija Ivanovska\corref{cor1}} 
\ead{marija.ivanovska@fe.uni-lj.si}
\author{Vitomir \v{S}truc}
\ead{vitomir.struc@fe.uni-lj.si}
\tnotetext[t1]{This manuscript is currently under review at Pattern Recognition Letters.}
\cortext[cor1]{Corresponding author}

\affiliation{organization={Faculty of Electrical Engineering, University of Ljubljana},
            addressline={Tr\v{z}a\v{s}ka cesta 25}, 
            city={Ljubljana},
            postcode={1000},
            country={Slovenia}}

\begin{abstract}
Face morphing attacks threaten biometric verification, yet most morphing attack detection (MAD) systems require task-specific training and generalize poorly to unseen attack types. Meanwhile, open-source multimodal large language models (MLLMs) have demonstrated strong visual–linguistic reasoning, but their potential in biometric forensics remains underexplored. In this paper, we present the first systematic zero-shot evaluation of open-source MLLMs for single-image MAD, using publicly available weights and a standardized, reproducible protocol. Across diverse morphing techniques, many MLLMs show non-trivial discriminative ability without any fine-tuning or domain adaptation, and LLaVA1.6-Mistral-7B achieves state-of-the-art performance, surpassing highly competitive task-specific MAD baselines by at least $23\%$ in terms of equal error rate (EER). The results indicate that multimodal pretraining can implicitly encode fine-grained facial inconsistencies indicative of morphing artifacts, enabling zero-shot forensic sensitivity. Our findings position open-source MLLMs as reproducible, interpretable, and competitive foundations for biometric security and forensic image analysis. This emergent capability also highlights new opportunities to develop state-of-the-art MAD systems through targeted fine-tuning or lightweight adaptation, further improving accuracy and efficiency while preserving interpretability. To support future research, all code and evaluation protocols will be released upon publication.
\end{abstract}



\begin{keyword}
Multimodal Large Language Models (MLLMs) \sep Morphing Attack Detection (MAD) \sep Zero-Shot Learning \sep Biometric Forensics \sep Face Recognition Security
\end{keyword}

\end{frontmatter}


\section{Introduction}
\vspace{-2px}
Face morphing attacks undermine the integrity of biometric verification by enabling multiple identities to be authenticated with a single morphed image. As morphs become increasingly realistic with minimal visible artifacts, existing morphing attack detection (MAD) methods often fail to recognize unseen attack types and typically lack interpretability, reducing their trustworthiness in security-critical applications~\cite{SYN_MAD_2022}. Meanwhile, recent advances in multi-modal large language models (MLLMs) have demonstrated emergent reasoning capacities through the alignment of visual and textual representations~\cite{hatef_survey_tifs_2025}. Their ability to link semantic understanding with fine-grained visual cues suggests that these models may inherently capture image inconsistencies, making them promising candidates for detecting subtle manipulation artifacts~\cite{ross_benchmark_2025, vishal_facexbench_2025}.\\
Despite their success in vision and forensic tasks, MLLMs remain largely unexplored for morphing attack detection. Most biometric security research still relies on task-specific, vision-only networks, while multimodal foundation models have received limited attention. Although CLIP-based setups have been studied before~\cite{Caldeira_MADation_2025_WACV, Raghu_2024_CVIP_CLIPzeroShot}, modern MLLMs surpass early encoders through instruction tuning and cross-modal reasoning, enabling deeper visual–semantic understanding. While some recent works have explored closed-source systems such as ChatGPT~\cite{zhang_chatGPT_MAD_2025,jia_gpt_detects_deepfakes_CVPRW_2024} for face forgery detection, their proprietary nature limits reproducibility and prevents transparent evaluation.\\
\begin{figure*}[!ht]
    \begin{center}
        \includegraphics[width=1\linewidth]{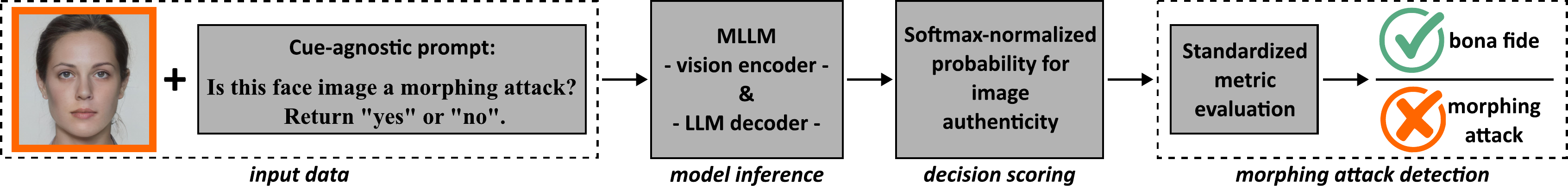}
        \vspace{-15px}
        \caption{\textbf{Method overview.} We evaluate MLLMs using standardized prompts and metric-based analysis to assess their inherent ability to detect manipulations.}\label{fig:overview}
        \vspace{-15px}
    \end{center}
\end{figure*}
In this work, we present the first systematic analysis of open-source MLLMs for single-image morphing attack detection (S-MAD). We focus on open models, as they enable standardized comparison, reproducibility, and community-driven benchmarking, while also allowing adaptation and fine-tuning for practical deployment in biometric forensics. We evaluate a broad set of recent MLLMs on standard benchmarks spanning diverse morphing techniques, using a strict zero-shot protocol to isolate each model’s inherent ability to reason about facial authenticity. We show that, despite being trained for general multimodal reasoning, many MLLMs exhibit non-trivial morph detection capability, and in some cases even surpass specialized state-of-the-art MAD systems explicitly trained for this task. These findings indicate that large-scale vision–language alignment encodes transferable perceptual priors that can detect morphological inconsistencies without task-specific supervision.\\
We make the following contributions: \textit{i)} a systematic zero-shot benchmark of open-source MLLMs for single-image morphing attack detection (S-MAD) across diverse datasets, using a standardized and reproducible protocol; \textit{ii)} a comprehensive analysis of inter-model performance patterns in evaluated MLLMs; \textit{iii)} a comparison against top-performing task-specific MAD systems, identifying LLaVA1.6-Mistral-7B as a new state-of-the-art open model that surpasses existing MADs; and \textit{iv)} insights into the emergent visual–semantic reasoning of MLLMs and its implications for generalization, interpretability, and the future of task supervision in biometric security.

\vspace{-7px}
\section{Related work}\label{sec:related_work}
\vspace{-2px}
This section first reviews recent open-source vision–language models, followed by a summary of research on single-image morphing attack detection (S-MAD) and related multimodal forensic studies.

\textbf{Multi-Modal Large Language Models.} Large Language Models (LLMs) have evolved from task-specific architectures into general-purpose systems capable of contextual reasoning and instruction following. Multi-modal Large Language Models (MLLMs) extend this capability by integrating visual encoders, grounding linguistic concepts in visual input.\\
Foundational models such as CLIP~\cite{Radford_2021_icml_CLIP}, Flamingo~\cite{jean_flamingo_NIPS_2022}, and BLIP~\cite{junnan_blip_ICML_2022} established joint image–text reasoning through contrastive and cross-attentive pretraining. Recent vision–language models—including LLaVA~\cite{Liu_LLaVA_NIPS_2023}, Qwen-VL~\cite{bai_QwenVL_2023}, InternVL~\cite{chen_InternVL_2024}, DeepSeek-VL2~\cite{wu_deepseek_2024}, Gemma~\cite{gemmateam_Gemma_2024}, and Pixtral~\cite{agrawal_Pixtral_2024} advance this foundation via instruction tuning, scalable fusion, and efficient attention mechanisms, achieving strong generalization in multi-modal reasoning. Their ability to align fine-grained visual features with semantic understanding makes them particularly promising for tasks such as forgery detection, where adaptability and interpretability are critical.

\textbf{Vision-Only Morphing Attack Detection.} Recent MAD approaches predominantly rely on deep learning~\cite{SYN_MAD_2022}. Damer \textit{et al.}~\cite{Naser_PW_MAD_2021} trained a pixel-wise CNN combining local and global supervision. Aghdaie \textit{et al.}~\cite{aaw_wvu_IJCB_2021} introduced an attention-aware wavelet network that enhances artifact-prone regions, and Borghi \textit{et al.} proposed UBO-SMAD-R3~\cite{ChiMo_dataset_maltoni_Access_2023}, a deeper Inception–ResNet variant trained on a large, morph-diverse dataset. Although these supervised systems achieve high in-domain accuracy, their generalization to unseen morphing techniques remains limited due to reliance on morph-specific training data. To mitigate this, several unsupervised approaches detect morphing artifacts without explicit labels: Fu \textit{et al.}~\cite{damer_iqa_IETB_2022} explored image-quality estimation, Fang \textit{et al.}~\cite{Damer2022_OC_MAD_SPA} applied reconstruction-based autoencoding, and Ivanovska \textit{et al.}~\cite{ivanovska2023mad_ddpm} advanced this line with diffusion-based modeling. Other work by Zhang~\textit{et al.} employed pretrained backbones for zero-shot detection~\cite{Zhang_2024_CVPRW_ViT}. Finally, self-supervised learning has emerged as a middle ground, with Ivanovska \textit{et al.}~\cite{selfMAD_FG_2025} showing that training models on synthetically simulated morph artifacts improves accuracy and robustness without labeled morph data.

\textbf{Vision-Language Models for Face Forgeries.} MLLMs have recently been explored for detecting face forgeries such as deepfakes and presentation attacks, with surveys highlighting their multimodal capabilities~\cite{hatef_survey_tifs_2025, ross_benchmark_2025}. Shi \textit{et al.}~\cite{shi_shield_VisIntell_2025} evaluated MLLMs for anti-spoofing and AI-generated image detection using multi-attribute CoT prompting under zero- and few-shot settings, finding promising but inconsistent robustness across datasets. Narayan \textit{et al.}~\cite{vishal_facexbench_2025} likewise showed that state-of-the-art MLLMs still struggle with fine-grained facial analysis despite strong general reasoning skills. Prompt-engineering studies further suggest that carefully crafted prompts can surface useful forensic cues without task-specific training~\cite{jia_gpt_detects_deepfakes_CVPRW_2024, ren_mllm_deepfake_detection_2025}.
For morphing attack detection (MAD), a few works have explored vision–language models in zero-shot settings. Zhang \textit{et al.}~\cite{zhang_chatGPT_MAD_2025} tested ChatGPT on print–scan morphs with handcrafted prompts, while Patwardhan \textit{et al.}~\cite{Raghu_2024_CVIP_CLIPzeroShot} used CLIP for interpretable S-MAD, comparing short versus long prompts and retrieving human-readable evidence. Caldeira \textit{et al.}~\cite{Caldeira_MADation_2025_WACV} proposed MADation, adapting CLIP for S-MAD and achieving competitive performance across benchmarks.

Despite progress in both vision-only and vision-language approaches, the potential of modern open-source MLLMs for morphing attack detection remains underexplored. Existing studies either rely on supervised methods or use closed-source models, limiting reproducibility and adaptability. Our work fills this gap by presenting the first systematic zero-shot benchmark of open-source MLLMs for single-image MAD, revealing their latent forensic sensitivity and showing that some models even surpass task-specific detectors while offering transparency, interpretability, and adaptability.

\vspace{-7px}
\section{Methodology}\label{sec:methodology}
We formulate single-image morphing attack detection (S-MAD) as a zero-shot visual reasoning problem, where a Multi-modal Large Language Model (MLLM) must infer the authenticity of a single input face image without task-specific supervision. The central objective is to determine whether vision–language pretraining confers latent sensitivity to morphing artifacts. Let $\mathcal{X}$ denote the space of facial images and $\mathcal{Y} = \{\texttt{yes}, \texttt{no}\}$ the binary label space, where $\texttt{yes}$ corresponds to a morphing attack and $\texttt{no}$ to a bona fide image. Each MLLM defines a conditional probability distribution
\begin{equation}
    P_\theta(y \mid x, p),
\end{equation}
parameterized by $\theta$ and conditioned on the textual prompt $p$. The decision rule
\begin{equation}
    \hat{y} = \arg\max_{y \in \mathcal{Y}} P_\theta(y \mid x, p)
\end{equation}
assigns the most probable label for an image $x$ and prompt $p$. 

Large language models (LLMs) are autoregressive, and generate text one token at a time. At each step, the model predicts a probability distribution over the next possible token given all previous tokens. Formally, if the current input sequence (the input prompt $p$ plus any previous tokens) is $t_1, t_2, \ldots, t_n$, the model computes a vector of logits $\mathbf{z}_n$ and derives
\begin{equation}
    P(t_{n+1} \mid t_1, \ldots, t_n) = \text{softmax}(\mathbf{z}_n).
\end{equation}
This quantity expresses how likely each word or symbol from the model's vocabulary is to appear next and is known as next-token probability estimation. In our experimental settings, we interpret this distribution as the model’s belief over the two possible labels, \texttt{"yes"} and \texttt{"no"}. For a given image~$x$ and prompt~$p$, the model produces the final-token logit vector $\mathbf{z}$ corresponding to the vocabulary distribution in the decoder. We extract the logits associated with tokens representing \texttt{"yes"} and \texttt{"no"} and apply the softmax function:
\begin{equation}\label{eq:decision_prob}
    P_\theta(y \mid x,p) = \frac{\exp(z_y)}{\exp(z_{\texttt{yes}})+\exp(z_{\texttt{no}})}, y \in \{\texttt{yes},\texttt{no}\}.
\end{equation}

The resulting $P_\theta(\texttt{yes}\mid x,p)$ expresses the probability that the model considers the input image to be a morphing attack, while $P_\theta(\texttt{no}\mid x,p)=1-P_\theta(\texttt{yes}\mid x,p)$ expresses the probability of the input image to be a bona fide.

To ensure consistent model behavior, each MLLM is prompted identically and assessed across multiple morphing datasets to measure its inherent discriminative ability without any task-specific adaptation. In our experiments we therefore employed the following fixed binary classification prompt designed for unambiguous outputs:

\begin{quote}
\small
\texttt{Is this face image a morphing attack? Return exactly one json on a single line (no code fences): \{"label":"yes"\} or \{"label":"no"\}. "yes" = morph, "no" = bona fide. No extra text.}
\end{quote}

This prompt constrains linguistic variability and enforces a uniform decision context across models. Importantly, it is cue-agnostic, as it does not reference any visual artifacts or facial regions, 
ensuring that the model’s responses rely solely on its internal visual–language reasoning rather than on handcrafted forensic hints. Each response was automatically parsed to extract the predicted label (\texttt{"yes"} or \texttt{"no"}). To quantify decision confidence, the probability of each label was estimated from the softmax-normalized logits returned by the model’s language decoder, as defined in Eq.~\ref{eq:decision_prob}. These normalized probabilities served as a continuous decision score that allows threshold-based computation of standardized, widely-used evaluation metrics. Overview of our method is showed in Fig.~\ref{fig:overview}.

\vspace{-7px}
\section{Experimental Setup}\label{sec:experimental_setup}
This section describes the model selection, evaluation datasets, metrics, and implementation details.

\textbf{Model Evaluation Details.} We evaluated nineteen open-source MLLMs spanning a range of architectures, training strategies, and parameter scales: InternVL3.5 (1B, 2B, 4B, 8B, 14B), DeepSeek-VL2 (Tiny, Small, VL2), Qwen2.5-VL (3B, 7B, 32B) Gemma3 (4B, 12B, 27B), LLaVA1.6 (Mistral-7B, Vicuna-7B, Vicuna-13B, 34B), and Pixtral-12B. All models and accompanying pretrained weights were obtained from the Hugging Face Model Hub\footnote{https://huggingface.co/docs/hub/en/models-the-hub} and used without any fine-tuning or adaptation. This diverse selection encompasses major open-source multi-modal frameworks based on CLIP-style vision encoders and transformer-based language backbones, allowing for a comparison across scales and design families. To ensure fair model comparison, and deterministic and reproducible outputs, generation was performed without stochastic sampling. Experiments were conducted on a single NVIDIA A100-SXM4 GPU with $80$\,GB of memory using Python 3.10, PyTorch 2.0, and CUDA 11.8. All code, prompts, and evaluation scripts will be released upon publication of the paper to support full reproducibility of the results.

\begin{table*}[t!]
\centering
\small
\caption{\textbf{Benchmarking Open-Source MLLMs.} Performance is measured in EER (\%) and BSCER@MACER(5\%). Medium-sized models, particularly LLaVA1.6-Mistral-7B, achieve the best overall accuracy, confirming that optimal forensic sensitivity does not necessarily scale with model size. [\textcolor{winner}{winner}, \textcolor{runner-up}{runner-up}]}
\label{tab:eer:prompt1}
\resizebox{\linewidth}{!}{%
\begin{tabular}{| l | c |  c |  c |  c |  c |  c |  c |  c |  c | c | c | c | c | c | c | c | c | c || c | c|}
\hline
\multirow{3}{*}{\textbf{Model}} & \multicolumn{10}{c|}{\textbf{FRLL-Morphs}} & \multicolumn{2}{c|}{\textbf{MIPGAN}} & \multicolumn{2}{c|}{\multirow{2}{*}{\textbf{MorDIFF}}} & \multicolumn{2}{c|}{\textbf{Morph-}} & \multicolumn{2}{c||}{\textbf{Greedy-}} & \multicolumn{2}{c|}{\multirow{2}{*}{\textbf{Average}}} \\ \cline{2-11}
 & \multicolumn{2}{c|}{\textbf{AMSL}} & \multicolumn{2}{c|}{\textbf{FM}$^*$} & \multicolumn{2}{c|}{\textbf{OCV}$^*$} & \multicolumn{2}{c|}{\textbf{SG}$^*$} & \multicolumn{2}{c|}{\textbf{WM}$^*$} & \multicolumn{2}{c|}{\textbf{II}} & \multicolumn{2}{c|}{} & \multicolumn{2}{c|}{\textbf{PIPE}} & \multicolumn{2}{c||}{\textbf{DiM}} & \multicolumn{1}{c}{} & \\ \cline {2-21}
 & \hlg{EER} & 5\% & \hlg{EER} & 5\% & \hlg{EER} & 5\% & \hlg{EER} & 5\% & \hlg{EER} & 5\% & \hlg{EER} & 5\% & \hlg{EER} & 5\% & \hlg{EER} & 5\% & \hlg{EER} & 5\% & \hlg{EER} & 5\% \\
\hline
\multicolumn{21}{c}{\textbf{Small MLLMs ($\mathbf{<}$ 7B parameters)}} \\
\hline
InternVL3.5-1B & \hlg{$14.71$} & $31.49$ & \hlg{$5.39$} & $6.96$ & \hlg{$1.96$} & $0.49$ & \hlg{$22.55$} & $70.28$ & \hlg{$10.29$} & $19.46$ & \hlg{$55.45$} & $97.57$ & \hlg{$32.84$} & $81.60$ & \hlg{$10.50$} & $23.07$ & \hlg{$40.69$} & $87.00$ & \hlg{$21.60$} & $46.44$\\
InternVL3.5-2B & \hlg{$25.00$} & $71.77$ & \hlg{$17.65$} & $29.62$ & \hlg{$20.59$} & $34.56$ & \hlg{$26.47$} & $53.41$ & \hlg{$16.18$} & $31.94$ & \hlg{$53.75$} & $99.24$ & \hlg{$9.80$} & $13.00$ & \hlg{$91.83$} & $99.92$ & \hlg{$18.63$} & $37.00$ & \hlg{$31.10$} & $52.27$\\
DeepSeek-VL2-Tiny & \hlg{$36.76$} & $69.20$ & \hlg{$1.47$} & $0.65$ & \hlg{$0.98$} & $0.41$ & \hlg{$\textcolor{winner}{\mathbf{2.94}}$} & $\textcolor{winner}{\mathbf{1.96}}$ & \hlg{$4.41$} & $4.50$ & \hlg{$10.98$} & $20.64$ & \hlg{$9.80$} & $20.40$ & \hlg{$\textcolor{winner}{\mathbf{0.27}}$} & $\textcolor{winner}{\mathbf{0.00}}$ & \hlg{$14.22$} & $36.00$ & \hlg{$9.09$} & $17.08$\\
Qwen2.5-VL-3B & \hlg{$55.39$} & $99.13$ & \hlg{$50.49$} & $95.42$ & \hlg{$50.49$} & $95.82$ & \hlg{$59.80$} & $100.00$ & \hlg{$35.29$} & $84.52$ & \hlg{$9.64$} & $26.86$ & \hlg{$22.06$} & $71.40$ & \hlg{$18.27$} & $35.74$ & \hlg{$33.82$} & $92.80$ & \hlg{$37.25$} & $77.97$\\
Gemma3-4B & \hlg{$40.20$} & $93.10$ & \hlg{$16.67$} & $34.29$ & \hlg{$11.76$} & $21.62$ & \hlg{$51.47$} & $99.10$ & \hlg{$24.02$} & $63.14$ & \hlg{$45.85$} & $88.16$ & \hlg{$16.18$} & $39.80$ & \hlg{$48.06$} & $93.47$ & \hlg{$27.45$} & $72.60$ & \hlg{$31.30$} & $67.25$\\
InternVL3.5-4B & \hlg{$22.06$} & $62.15$ & \hlg{$9.80$} & $13.09$ & \hlg{$3.92$} & $3.36$ & \hlg{$36.76$} & $89.20$ & \hlg{$11.76$} & $20.39$ & \hlg{$55.56$} & $95.60$ & \hlg{$21.08$} & $37.60$ & \hlg{$16.00$} & $41.50$ & \hlg{$34.31$} & $65.60$ & \hlg{$23.47$} & $47.61$\\
\hline
\multicolumn{21}{c}{\textbf{Medium MLLMs (7B-17B parameters)}} \\
\hline
LLaVa1.6-Mistral-7B & \hlg{$\textcolor{winner}{\mathbf{0.98}}$} & $\textcolor{winner}{\mathbf{0.00}}$ & \hlg{$\textcolor{winner}{\mathbf{0.00}}$} & $\textcolor{winner}{\mathbf{0.00}}$ & \hlg{$\textcolor{winner}{\mathbf{0.00}}$} & $\textcolor{winner}{\mathbf{0.00}}$ & \hlg{$\textcolor{runner-up}{\mathbf{5.39}}$} & $16.20$ & \hlg{$\textcolor{winner}{\mathbf{0.00}}$} & $\textcolor{winner}{\mathbf{0.00}}$ & \hlg{$\textcolor{winner}{\mathbf{2.21}}$} & $\textcolor{runner-up}{\mathbf{6.24}}$ & \hlg{$8.33$} & $\textcolor{runner-up}{\mathbf{5.40}}$ & \hlg{$\textcolor{runner-up}{\mathbf{0.47}}$} & $\textcolor{winner}{\mathbf{0.00}}$ & \hlg{$\textcolor{winner}{\mathbf{7.35}}$} & $37.80$ & \hlg{$\textcolor{winner}{\mathbf{2.75}}$} & $\textcolor{winner}{\mathbf{7.29}}$\\
LLaVa1.6-Vicuna-7B & \hlg{$6.86$} & $8.64$ & \hlg{$\textcolor{runner-up}{\mathbf{0.49}}$} & $\textcolor{runner-up}{\mathbf{0.08}}$ & \hlg{$\textcolor{winner}{\mathbf{0.00}}$} & $\textcolor{winner}{\mathbf{0.00}}$ & \hlg{$6.86$} & $\textcolor{runner-up}{\mathbf{11.95}}$ & \hlg{$\textcolor{runner-up}{\mathbf{0.49}}$} & $0.49$ & \hlg{$22.16$} & $62.67$ & \hlg{$15.69$} & $35.00$ & \hlg{$40.16$} & $97.65$ & \hlg{$9.80$} & $40.20$ & \hlg{$11.39$} & $28.52$\\
Qwen2.5-VL-7B & \hlg{$11.76$} & $27.17$ & \hlg{$7.35$} & $12.03$ & \hlg{$9.31$} & $14.58$ & \hlg{$11.27$} & $22.91$ & \hlg{$5.88$} & $6.22$ & \hlg{$25.17$} & $61.31$ & \hlg{$16.67$} & $27.80$ & \hlg{$15.39$} & $45.07$ & \hlg{$25.00$} & $54.60$ & \hlg{$14.20$} & $30.19$\\
InternVL3.5-8B & \hlg{$10.78$} & $21.15$ & \hlg{$1.96$} & $1.06$ & \hlg{$1.96$} & $0.49$ & \hlg{$25.98$} & $50.16$ & \hlg{$5.39$} & $6.14$ & \hlg{$58.10$} & $97.57$ & \hlg{$\textcolor{winner}{\mathbf{0.49}}$} & $\textcolor{winner}{\mathbf{0.00}}$ & \hlg{$74.50$} & $99.01$ & \hlg{$49.51$} & $91.60$ & \hlg{$25.41$} & $40.80$\\
Gemma3-12B & \hlg{$39.22$} & $95.03$ & \hlg{$17.16$} & $29.95$ & \hlg{$9.80$} & $11.79$ & \hlg{$17.65$} & $48.53$ & \hlg{$22.55$} & $46.36$ & \hlg{$38.09$} & $81.34$ & \hlg{$5.39$} & $6.60$ & \hlg{$5.96$} & $8.12$ & \hlg{$12.25$} & $\textcolor{runner-up}{\mathbf{28.20}}$ & \hlg{$18.67$} & $39.55$\\
Pixtral-12B & \hlg{$\textcolor{runner-up}{\mathbf{1.96}}$} & $\textcolor{runner-up}{\mathbf{0.41}}$ & \hlg{$\textcolor{runner-up}{\mathbf{0.49}}$} & $\textcolor{runner-up}{\mathbf{0.08}}$ & \hlg{$4.41$} & $3.52$ & \hlg{$11.76$} & $29.46$ & \hlg{$1.96$} & $0.49$ & \hlg{$41.83$} & $84.22$ & \hlg{$13.73$} & $36.00$ & \hlg{$3.75$} & $\textcolor{runner-up}{\mathbf{3.26}}$ & \hlg{$28.43$} & $82.80$ & \hlg{$12.04$} & $26.69$\\
LLaVa1.6-Vicuna-13B & \hlg{$8.33$} & $18.94$ & \hlg{$\textcolor{runner-up}{\mathbf{0.49}}$} & $\textcolor{runner-up}{\mathbf{0.08}}$ & \hlg{$\textcolor{runner-up}{\mathbf{0.49}}$} & $0.00$ & \hlg{$9.31$} & $25.70$ & \hlg{$2.94$} & $1.72$ & \hlg{$\textcolor{runner-up}{\mathbf{4.95}}$} & $9.03$ & \hlg{$\textcolor{runner-up}{\mathbf{3.43}}$} & $7.60$ & \hlg{$21.75$} & $58.65$ & \hlg{$\textcolor{runner-up}{\mathbf{9.31}}$} & $\textcolor{winner}{\mathbf{22.60}}$ & \hlg{$\textcolor{runner-up}{\mathbf{6.78}}$} & $\textcolor{runner-up}{\mathbf{16.04}}$\\
InternVL3.5-14B & \hlg{$87.75$} & $97.56$ & \hlg{$25.98$} & $37.64$ & \hlg{$15.20$} & $20.15$ & \hlg{$58.82$} & $97.87$ & \hlg{$41.18$} & $50.70$ & \hlg{$58.43$} & $97.04$ & \hlg{$33.33$} & $38.80$ & \hlg{$7.76$} & $11.53$ & \hlg{$39.22$} & $68.20$ & \hlg{$40.85$} & $57.72$\\
DeepSeek-VL2-Small & \hlg{$3.43$} & $1.56$ & \hlg{$\textcolor{runner-up}{\mathbf{0.49}}$} & $0.16$ & \hlg{$\textcolor{runner-up}{\mathbf{0.49}}$} & $0.25$ & \hlg{$11.76$} & $28.40$ & \hlg{$\textcolor{runner-up}{\mathbf{0.49}}$} & $\textcolor{runner-up}{\mathbf{0.16}}$ & \hlg{$32.80$} & $78.76$ & \hlg{$5.39$} & $6.20$ & \hlg{$4.02$} & $3.34$ & \hlg{$16.67$} & $40.80$ & \hlg{$8.39$} & $17.74$\\
\hline
\multicolumn{21}{c}{\textbf{Large MLLMs ($\mathbf{>}$ 17B parameters)}} \\
\hline
Gemma3-27B & \hlg{$30.39$} & $91.68$ & \hlg{$18.14$} & $41.33$ & \hlg{$9.80$} & $19.25$ & \hlg{$30.39$} & $98.69$ & \hlg{$14.71$} & $46.36$ & \hlg{$51.67$} & $96.43$ & \hlg{$23.04$} & $49.80$ & \hlg{$26.64$} & $70.71$ & \hlg{$19.12$} & $57.00$ & \hlg{$24.88$} & $63.47$\\
DeepSeek-VL2 & \hlg{$4.41$} & $4.00$ & \hlg{$0.98$} & $0.25$ & \hlg{$1.47$} & $0.66$ & \hlg{$30.88$} & $91.98$ & \hlg{$0.98$} & $0.08$ & \hlg{$43.78$} & $89.61$ & \hlg{$32.84$} & $84.00$ & \hlg{$35.61$} & $80.35$ & \hlg{$15.20$} & $30.00$ & \hlg{$18.46$} & $42.33$\\
Qwen2.5-VL-32B & \hlg{$33.33$} & $78.53$ & \hlg{$15.69$} & $33.55$ & \hlg{$15.20$} & $32.19$ & \hlg{$12.25$} & $24.80$ & \hlg{$20.59$} & $43.16$ & \hlg{$5.22$} & $\textcolor{winner}{\mathbf{5.31}}$ & \hlg{$26.47$} & $89.20$ & \hlg{$23.76$} & $37.86$ & \hlg{$12.75$} & $35.20$ & \hlg{$18.36$} & $42.20$\\
LLaVa1.6-34B & \hlg{$16.18$} & $51.77$ & \hlg{$\textcolor{runner-up}{\mathbf{0.49}}$} & $\textcolor{runner-up}{\mathbf{0.08}}$ & \hlg{$\textcolor{winner}{\mathbf{0.00}}$} & $\textcolor{runner-up}{\mathbf{0.08}}$ & \hlg{$7.84$} & $13.01$ & \hlg{$3.92$} & $3.44$ & \hlg{$18.41$} & $33.31$ & \hlg{$15.20$} & $35.80$ & \hlg{$46.39$} & $84.45$ & \hlg{$18.14$} & $45.60$ & \hlg{$14.06$} & $29.73$\\
\hline
\hline
\textbf{Average} & \hlg{$23.66$} & $48.59$ & \hlg{$\textcolor{runner-up}{\mathbf{10.06}}$} & $\textcolor{runner-up}{\mathbf{17.70}}$ & \hlg{$\textcolor{winner}{\mathbf{8.31}}$} & $\textcolor{runner-up}{\mathbf{13.64}}$ & \hlg{$23.17$} & $51.24$ & \hlg{$11.74$} & $22.59$ & \hlg{$33.37$} & $65.31$ & \hlg{$16.41$} & $38.74$ & \hlg{$25.85$} & $47.04$ & \hlg{$22.73$} & $54.51$ & \hlg{$19.48$} & $39.93$\\ \hline
\multicolumn{21}{l}{$^*$FM: FaceMorpher, OCV: OpenCV, SG: StyleGAN2, WM: WebMorph;}
\vspace{-15px}
\end{tabular}}
\end{table*}
\textbf{Evaluation Metrics.} To calculate the decision scoring we used the model's final-step decoder logits, as defined in Eq.~\ref{eq:decision_prob}. The tokenizers may segment leading spaces or capitalization differently, so we aggregated logits over token variants for each label (e.g.,``\texttt{yes}'', ``\texttt{ yes}'', ``\texttt{Yes}'', ``\texttt{ Yes}'') and likewise for ``\texttt{no}''. When both variant sets are empty, we set $P_\theta(\texttt{yes}\mid x,p)=0.5$. Computed probabilities served as continuous scores for computing widely-used, standardized metrics in line with ISO/IEC 20059:2025 standards\footnote{International Organization for Standardization (ISO). ISO/IEC 20059:2025 — Information technology — Biometric presentation attack detection — Testing and reporting. (This standard supersedes ISO/IEC 30107-3:2017.)}. Specifically, we calculate the detection Equal Error Rate (EER), where the Morphing Attack Classification Error Rate (MACER), equals the Bona Fide Sample Classification Error Rate (BSCER). MACER measures the proportion of morphing attacks incorrectly accepted as bona fide, while BSCER reflects the proportion of bona fide images incorrectly rejected as attacks. In addition to EER, we report BSCER values at fixed MACER operating point of $5\%$, to provide a more realistic assessment of morphing attack detection performance in high-security scenarios and better reflect the system’s robustness under low false-acceptance conditions.

\textbf{Testing Data.} Evaluation was conducted on five widely-used single-image morphing datasets representing different morphing generation techniques, i.e. FRLL-Morphs~\cite{sarkar_frll_frgc_feret_morphs_2020}, MIPGAN II~\cite{Zhang_MIPGAN_2021}, MorDIFF~\cite{damer_MorDIFF_IWBF_2023}, Morph-PIPE~\cite{Zhang_PIPE_IJCB_2024}, and Greedy-DiM~\cite{Blasingame_greedy_IJCB_2024}. Although retired, FRLL-Morphs is used in our evaluations to ensure fair comparison with previously published methods~\cite{Caldeira_MADation_2025_WACV, selfMAD_FG_2025, Damer2022_OC_MAD_SPA} that follow the same established evaluation protocols. This dataset was obtained prior to its withdrawal. Nevertheless its generation procedures are publicly documented, allowing replication following the generation steps detailed in the corresponding dataset paper. The rest of the evaluation datasets are publicly available and together with FRLL-Morphs capture a broad spectrum of morphing characteristics, from traditional landmark-based morphs to generative, GAN- and diffusion-based manipulations. We used Dlib's face detector to detect faces in each testing image and cropped the face out with a margin of $12.5\%$, following protocols published in previous works~\cite{selfMAD_FG_2025}. The rest of the preprocessing, such as image resizing and normalization, is by default implemented inside each evaluated model's pipeline.
\vspace{-5px}
\section{Results}\label{Sec: Results}
\textbf{Benchmarking Open-Source MLLMs.} Morphing attack detection (MAD) performance of all evaluated open-source Multi-modal Large Language Models (MLLMs) is summarized in Tbl.~\ref{tab:eer:prompt1}. We divide the models into three subcategories based on their size: small models (fewer than 7 billion parameters), medium-sized models (between 7 and 17 billion parameters), and large models (above 17 billion parameters). The results reveal pronounced variability in zero-shot MAD capability among the models. In general, all models demonstrate meaningful discriminative ability, with EER values far below random guessing, despite being trained primarily for general-purpose visual reasoning. These results suggest that multimodal instruction-tuned models develop latent awareness of subtle facial inconsistencies (such as texture discontinuities, blending boundaries, or geometric asymmetries) as part of their generic visual reasoning process. However, we note that smaller architectures exhibited limited discriminative capacity, except for DeepSeek-VL2-Tiny, which achieves an impressive average EER of $9.09\%$ across various morphing techniques, effectively leveraging its Mixture-of-Experts (MoE) design to maintain strong multimodal alignment despite its compact size. Its medium-sized variant, DeepSeek-VL2-Small, only slightly improves the overall detection performance with an average EER of $8.39\%$.
\begin{figure}[!t]
    \begin{center}
        \includegraphics[width=1\linewidth]{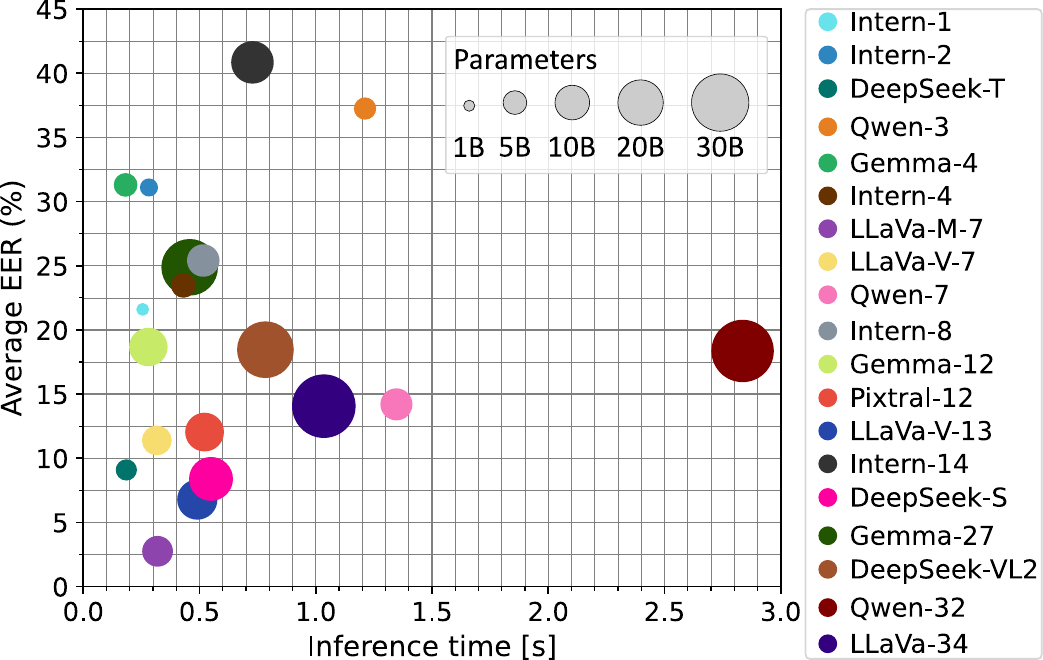}\vspace{-1mm}
        \caption{\textbf{Accuracy–efficiency trade-off of evaluated MLLMs.} Medium-sized models achieve the best balance, showing optimal forensic sensitivity.} \label{fig:eer_vs_time}
    \end{center}
\vspace{-25px}
\end{figure}
\begin{table*}[!t]
\centering
\small
\caption{\textbf{Best MLLM vs. Task-Specific MADs.} LLaVA1.6-Mistral-7B outperforms existing MADs, without task-specific training. [\textcolor{winner}{winner}, \textcolor{runner-up}{runner-up}]}
\label{tab:eer:vsMADs}
\resizebox{\linewidth}{!}{%
\begin{tabular}{|c | l | c | c | c | c | c | c | c | c | c | c | c | c | c | c | c | c | c | c || c | c |}
\hline
\multicolumn{2}{|c|}{\multirow{3}{*}{\textbf{Model}}} & \multicolumn{10}{c|}{\textbf{FRLL-Morphs}} & \multicolumn{2}{c|}{\textbf{MIPGAN}} & \multicolumn{2}{c|}{\multirow{2}{*}{\textbf{MorDIFF}}} & \multicolumn{2}{c|}{\textbf{Morph-}} & \multicolumn{2}{c||}{\textbf{Greedy-}} & \multicolumn{2}{c|}{\multirow{2}{*}{\textbf{Average}}} \\ \cline{3-12}
\multicolumn{1}{|c}{} & & \multicolumn{2}{c|}{\textbf{AMSL}} & \multicolumn{2}{c|}{\textbf{FM}$^{**}$} & \multicolumn{2}{c|}{\textbf{OCV}$^{**}$} & \multicolumn{2}{c|}{\textbf{SG}$^{**}$} & \multicolumn{2}{c|}{\textbf{WM}$^{**}$} & \multicolumn{2}{c|}{\textbf{II}} & \multicolumn{2}{c|}{} & \multicolumn{2}{c|}{\textbf{PIPE}} & \multicolumn{2}{c||}{\textbf{DiM}} & \multicolumn{2}{c|}{}\\ \cline{3-22}
\multicolumn{1}{|c}{} & & \hlg{EER} & 5\% & \hlg{EER} & 5\% & \hlg{EER} & 5\% & \hlg{EER} & 5\% & \hlg{EER} & 5\% & \hlg{EER} & 5\% & \hlg{EER} & 5\% & \hlg{EER} & 5\% & \hlg{EER} & 5\% & \hlg{EER} & 5\% \\
\hline\hline
\multirow{6}{*}{U$^*$} & FIQA-MagFace~\cite{damer_iqa_IETB_2022} & \hlg{$30.94$} & $77.94$ & \hlg{$27.99$} & $73.04$ & \hlg{$24.73$} & $66.18$ & \hlg{$\textcolor{runner-up}{\mathbf{7.53}}$} & $\textcolor{winner}{\mathbf{8.82}}$ & \hlg{$27.19$} & $68.14$ & \hlg{$24.51$} & $65.71$ & \hlg{$9.80$} & $22.55$ & \hlg{$49.62$} & $91.54$ & \hlg{$47.00$} & $94.61$ & \hlg{$27.70$} & $63.17$\\
& CNNIQA~\cite{damer_iqa_IETB_2022} & \hlg{$21.61$} & $60.29$ & \hlg{$19.97$} & $57.84$ & \hlg{$7.53$} & $11.76$ & \hlg{$35.92$} & $75.49$ & \hlg{$21.54$} & $46.57$ & \hlg{$23.44$} & $53.57$ & \hlg{$36.20$} & $89.22$ & \hlg{$66.54$} & $98.83$ & \hlg{$49.40$} & $96.08$ & \hlg{$31.35$} & $65.52$\\
& SPL-MAD~\cite{Damer2022_OC_MAD_SPA} & \hlg{$3.26$} & $0.50$ & \hlg{$1.03$} & $0.99$ & \hlg{$1.88$} & $0.50$ & \hlg{$14 .65$} & $32.18$ & \hlg{$6.39$} & $11.39$ & \hlg{$2.35$} & $1.45$ & \hlg{$9.78$} & $23.27$ & \hlg{$18.88$} & $33.62$ & \hlg{$37.72$} & $80.69$ & \hlg{$10.66$} & $20.51$\\
& MAD-DDPM~\cite{ivanovska2023mad_ddpm} & \hlg{$27.13$} & $94.94$ & \hlg{$10.40$} & $95.19$ & \hlg{$13.76$} & $95.17$ & \hlg{$14.32$} & $95.17$ & \hlg{$30.30$} & $95.09$ & \hlg{$4.70$} & $95.30$ & \hlg{$2.80$} & $96.40$ & \hlg{$13.88$} & $95.14$ & \hlg{$36.10$} & $95.20$ & \hlg{$17.04$} & $95.29$\\
& CLIP-ZSL~\cite{Raghu_2024_CVIP_CLIPzeroShot} & \hlg{$31.86$} & $62.11$ & \hlg{$3.92$} & $2.62$ & \hlg{$7.35$} & $7.78$ & \hlg{$17.65$} & $35.84$ & \hlg{$18.63$} & $27.76$ & \hlg{$\textcolor{winner}{\mathbf{0.00}}$} & $\textcolor{winner}{\mathbf{0.00}}$ & \hlg{$23.53$} & $36.40$ & \hlg{$25.36$} & $61.46$ & \hlg{$14.22$} & $\textcolor{runner-up}{\mathbf{23.80}}$ & \hlg{$15.84$} & $28.64$\\
& ViT~\cite{Zhang_2024_CVPRW_ViT} & \hlg{$21.08$} & $52.00$ & \hlg{$\textcolor{runner-up}{\mathbf{0.98}}$} & $0.33$ & \hlg{$\textcolor{runner-up}{\mathbf{0.98}}$} & $\textcolor{runner-up}{\mathbf{0.25}}$ & \hlg{$20.10$} & $46.81$ & \hlg{$13.73$} & $30.06$ & \hlg{$2.86$} & $1.67$ & \hlg{$\textcolor{runner-up}{\mathbf{1.47}}$} & $\textcolor{winner}{\mathbf{0.60}}$ & \hlg{$12.60$} & $32.17$ & \hlg{$49.02$} & $83.80$ & \hlg{$13.65$} & $27.52$\\ \hline
\multirow{5}{*}{S$^*$} & MixFaceNet~\cite{Boutros_MixFaceNet_2021} & \hlg{$31.03$} & $65.56$ & \hlg{$8.37$} & $11.51$ & \hlg{$9.85$} & $15.89$ & \hlg{$38.92$} & $84.94$ & \hlg{$31.03$} & $78.38$ & \hlg{$32.14$} & $71.62$ & \hlg{$5.88$} & $6.00$ & \hlg{$33.69$} & $85.66$ & \hlg{$39.71$} & $90.20$ & \hlg{$25.56$} & $56.64$\\
& PW-MAD~\cite{Naser_PW_MAD_2021} & \hlg{$4.43$} & $4.64$ & \hlg{$1.97$} & $1.72$ & \hlg{$2.46$} & $1.80$ & \hlg{$17.24$} & $34.62$ & \hlg{$9.85$} & $12.29$ & \hlg{$4.29$} & $3.95$ & \hlg{$11.27$} & $18.40$ & \hlg{$20.40$} & $55.39$ & \hlg{$42.16$} & $95.40$ & \hlg{$12.67$} & $25.36$\\
& MADation~\cite{Caldeira_MADation_2025_WACV} & \hlg{$27.94$} & $61.43$ & \hlg{$1.47$} & $0.74$ & \hlg{$1.47$} & $1.39$ & \hlg{$19.61$} & $32.57$ & \hlg{$24.51$} & $42.67$ & \hlg{$\textcolor{winner}{\mathbf{0.00}}$} & $\textcolor{runner-up}{\mathbf{0.15}}$ & \hlg{$24.02$} & $38.00$ & \hlg{$44.43$} & $89.30$ & \hlg{$30.88$} & $62.60$ & \hlg{$19.37$} & $36.54$\\
& AAW-MAD$^{\dagger}$~\cite{aaw_wvu_IJCB_2021} & \hlg{$1.48$} & $0.70$ & \hlg{$1.97$} & $2.23$ & \hlg{$11.82$} & $28.67$ & \hlg{$44.33$} & $90.10$ & \hlg{$\textcolor{runner-up}{\mathbf{2.46}}$} & $2.13$ & \hlg{$23.22$} & $95.07$ & \hlg{$13.24$} & $26.20$ & \hlg{$29.87$} & $95.30$ & \hlg{$\textcolor{winner}{\mathbf{1.47}}$} & $\textcolor{winner}{\mathbf{0.40}}$ & \hlg{$14.43$} & $37.87$\\
& UBO-R3$^{\ddagger}$~\cite{ChiMo_dataset_maltoni_Access_2023} & \hlg{$-$} & $-$ & \hlg{$-$} & $-$ & \hlg{$-$} & $-$ & \hlg{$-$} & $-$ & \hlg{$-$} & $-$ & \hlg{$2.86$} & $1.14$ & \hlg{$3.43$} & $\textcolor{runner-up}{\mathbf{1.60}}$ & \hlg{$\textcolor{runner-up}{\mathbf{4.70}}$} & $\textcolor{runner-up}{\mathbf{3.57}}$ & \hlg{$10.78$} & $24.80$ & \hlg{$5.44$} & $\textcolor{runner-up}{\mathbf{7.78}}$\\ \hline
SS$^*$ & SelfMAD~\cite{selfMAD_FG_2025} & \hlg{$\textcolor{runner-up}{\mathbf{0.99}}$} & $\textcolor{runner-up}{\mathbf{0.05}}$ & \hlg{$\textcolor{winner}{\mathbf{0.00}}$} & $\textcolor{runner-up}{\mathbf{0.26}}$ & \hlg{$\textcolor{winner}{\mathbf{0.00}}$} & $\textcolor{winner}{\mathbf{0.00}}$ & \hlg{$10.34$} & $24.22$ & \hlg{$3.45$} & $\textcolor{runner-up}{\mathbf{1.64}}$ & \hlg{$2.95$} & $1.97$ & \hlg{$\textcolor{winner}{\mathbf{0.99}}$} & $5.20$ & \hlg{$5.89$} & $12.44$ & \hlg{$7.60$} & $37.60$ & \hlg{$\textcolor{runner-up}{\mathbf{3.58}}$} & $9.26$\\ 
\hline\hline
\multicolumn{2}{|c|}{LLaVa1.6-Mistral-7B} & \hlg{$\textcolor{winner}{\mathbf{0.98}}$} & $\textcolor{winner}{\mathbf{0.00}}$ & \hlg{$\textcolor{winner}{\mathbf{0.00}}$} & $\textcolor{winner}{\mathbf{0.00}}$ & \hlg{$\textcolor{winner}{\mathbf{0.00}}$} & $\textcolor{winner}{\mathbf{0.00}}$ & \hlg{$\textcolor{winner}{\mathbf{5.39}}$} & $\textcolor{runner-up}{\mathbf{16.20}}$ & \hlg{$\textcolor{winner}{\mathbf{0.00}}$} & $\textcolor{winner}{\mathbf{0.00}}$ & \hlg{$\textcolor{runner-up}{\mathbf{2.21}}$} & $6.24$ & \hlg{$8.33$} & $5.40$ & \hlg{$\textcolor{winner}{\mathbf{0.47}}$} & $\textcolor{winner}{\mathbf{0.00}}$ & \hlg{$\textcolor{runner-up}{\mathbf{7.35}}$} & $37.80$ & \hlg{$\textcolor{winner}{\mathbf{2.75}}$} & $\textcolor{winner}{\mathbf{7.29}}$\\
\hline
\multicolumn{22}{l}{$^*$U: unsupervised, S: supervised, SS: self-supervised; $^{**}$FM: FaceMorpher, OCV: OpenCV, SG: StyleGAN2, WM: WebMorph;} \\
\multicolumn{22}{l}{$\dagger$ SOTA from the FVC-onGoing benchmarking platform; $\ddagger$ SOTA from the NIST FATE MORPH benchmarking platform;}
\vspace{-15px}
\end{tabular}}
\end{table*}
\begin{figure*}[!t]
    \begin{center}
        \includegraphics[width=1\linewidth]{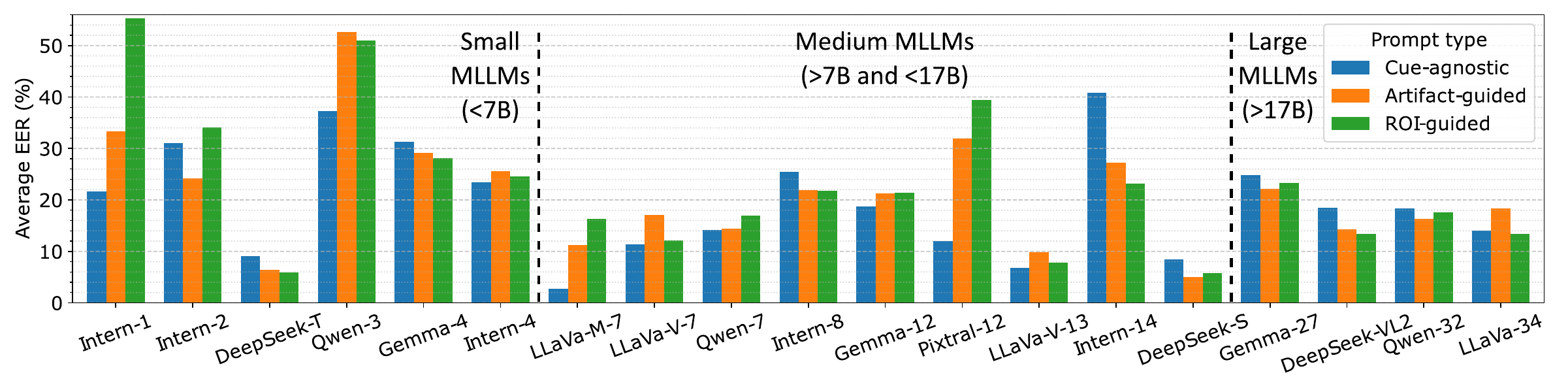}
        \vspace{-25px}
        \caption{\textbf{Prompt design impact.} Complex prompts degrade performance in small and medium models, while larger models benefit from richer semantic guidance.} \label{fig:eer_prompt_type}
    \end{center}
\vspace{-15px} \end{figure*}
Nevertheless, medium-sized MLLMs in general exhibit substantially stronger detection accuracy compared to small ones. Notably, LLaVA1.6-Mistral-7B achieves the lowest average EER of $2.75\%$, establishing a new state of the art among all considered models. Although larger variants (above 17 billion parameters) consistently outperform their smaller counterparts, they show degraded performance compared to medium-sized models, suggesting that model scale alone does not guarantee stronger forensic sensitivity. This trend is further illustrated in Fig.~\ref{fig:eer_vs_time}, which plots the average EER against the inference time per image, with marker size reflecting model scale. The figure reveals a clear efficiency–accuracy trade-off: medium-sized models offer the most favorable balance between detection accuracy and computational cost. We observe similar size–performance correlations when analyzing results across individual morphing techniques and datasets. In addition, we observe that MLLMs perform best on artifact-rich morphing datasets, such as the landmark-based FRLL-Morphs samples generated with FaceMorpher (FM), OpenCV (OCV), and WebMorph (WM). In most MLLMs, detection accuracy drops significantly on post-processed morphs where visible artifacts are minimized after morph generation (e.g., FRLL-Morphs AMSL). Similarly, relatively high EERs were observed for deep learning–based morphs, including the higher-quality GAN-based MIPGAN~II and the more recent Morph-PIPE and Greedy-DiM datasets.\\
\textbf{Comparison With Task-Specific MADs.} To contextualize the detection accuracy of our best-performing MLLM, we compare LLaVA1.6-Mistral-7B against top-performing morphing attack detectors (MADs) specifically developed and trained for this task. Tbl.~\ref{tab:eer:vsMADs} reports the results. LLaVA1.6-Mistral-7B outperforms all competitive methods representing diverse model designs and training objectives, improving upon the runner-up SelfMAD by $23\%$ in EER and $21\%$ in BSCER@MACER(5\%). Among the unsupervised baselines, we consider FIQA-MagFace, CNNIQA, SPL-MAD, MAD-DDPM, and the zero-shot methods CLIP-ZSL and ViT. Among them, SPL-MAD achieves the best performance with an average EER of $10.66\%$ and BSCER@MACER(5\%) of $20.51\%$, corresponding to a $74\%$ drop in detection accuracy compared to LLaVA1.6-Mistral-7B. Interestingly, CLIP-ZSL—a zero-shot method conceptually related to MLLMs—achieves an EER of $15.84\%$ and BSCER@MACER(5\%) of $28.64\%$, reflecting the limitations of CLIP’s vision–language alignment, which lacks the contextual reasoning and adaptive instruction-following capabilities of modern multimodal LLMs. We further compare against supervised MADs MixFaceNet, PW-MAD, and MADation, all trained on the large-scale synthetic SMDD dataset for optimal MAD accuracy and generalization. In addition, we include state-of-the-art supervised methods from two internationally recognized benchmarks: the Inception-based UBO-R3 from the FVC-onGoing platform and AAW-MAD from the NIST FATE MORPH benchmark. While these task-specific models benefit from explicit morph supervision, their performance remains largely dataset-dependent, revealing limited generalization capabilities. UBO-R3, the state-of-the-art in this category, achieves an EER of $5.44\%$ and BSCER@MACER(5\%) of $7.78\%$, representing nearly a $50\%$ accuracy drop compared to LLaVA1.6-Mistral-7B, despite being trained on a wide variety of morphing techniques including FRLL-Morphs samples. Notably, MADation, which fine-tunes the vision–language model CLIP for MAD, demonstrates relatively poor results with an EER of $19.37\%$ and BSCER@MACER(5\%) of $36.54\%$, highlighting the gap between conventional CLIP adaptation and the richer cross-modal reasoning capabilities learned by instruction-tuned MLLMs.\\
\begin{figure*}[!ht]
    \begin{center}
        \includegraphics[width=1\linewidth]{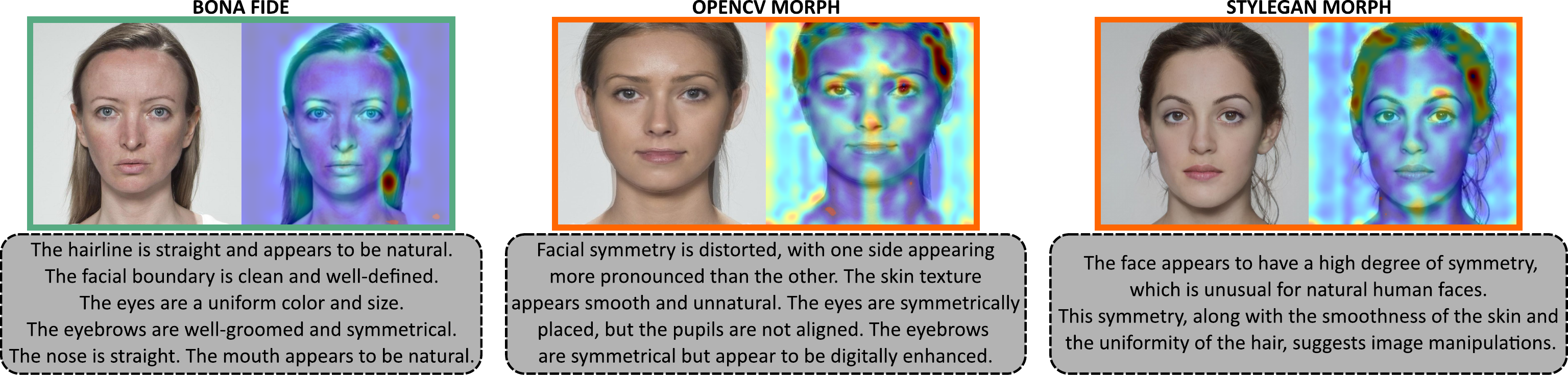}
        \vspace{-15px}
        \caption{\textbf{Interpretability Analysis.} LLaVA1.6-Mistral-7B's reasoning and attention highlight key facial cues such as symmetry, texture, and boundary consistency.}\label{fig:qualitative}
        \vspace{-15px}
    \end{center}
\end{figure*}
\textbf{Impact of Prompt Design.} To assess the influence of prompt design on MLLM performance, we re-evaluate all models using two additional prompts—an artifact-guided prompt and a region-of-interest (ROI)–guided prompt. The former explicitly directs the model to look for typical morphing artifacts such as ghosting, blended edges, or geometric asymmetries, while the latter instructs it to focus on specific facial areas—such as the eyes, nose, mouth, and facial boundary—where morphing artifacts most often occur, but without describing the artifacts that may be present. For each model, we compute the average EER across all evaluation datasets and visualize the results for all three prompts using a bar plot (Fig.~\ref{fig:eer_prompt_type}). The analysis reveals that small and medium-sized MLLMs generally exhibit degraded performance when the more complex artifact- and ROI-guided prompts are used. This behavior likely stems from their limited multimodal alignment depth and attention capacity, which prevents them from fully parsing and operationalizing such instructions—so the additional guidance introduces confusion rather than improvement. Moreover, these explicit prompts may overconstrain the reasoning process, causing the model to overlook unmentioned artifacts or facial regions and reducing sensitivity to unforeseen morphing patterns. In contrast, larger MLLMs tend to benefit from these structured prompts, displaying reduced EER in most cases. This indicates that higher-capacity models are better equipped to interpret and utilize semantically rich guidance. The observed trend reflects an emerging capacity in instruction-tuned MLLMs to integrate visual reasoning with semantic cues—a property central to their potential use in explainable biometric forensics.\\
\textbf{Visual–Linguistic Interpretability.} To further investigate interpretability, we prompted the best-performing model (LLaVA1.6-Mistral-7B) to describe the visual cues underlying its morphing-attack decisions and analyzed the corresponding attention maps of its vision tower. Fig.~\ref{fig:qualitative} illustrates three representative cases: a bona fide face, a landmark-based morph, and a GAN-based morph. The model’s verbal reasoning correlated strongly with the attention heatmaps, indicating coherent visual-semantic grounding. For bona fide images, the model focused on well-aligned structural cues such as clean facial boundaries, natural hairlines, and symmetric features. In contrast, morphing attacks elicited attention over artifact-prone areas, with explanations emphasizing distorted facial symmetry, irregular texture smoothness, and inconsistent hairline or boundary transitions. Overall, these findings suggest that LLaVA1.6-Mistral-7B’s decisions arise from semantically consistent visual reasoning, highlighting its potential for interpretable biometric forensics.
\vspace{-7px}
\section{Conclusion}
\vspace{-5px}
This work introduces a protocol and presents the first comprehensive zero-shot benchmark for evaluating open-source Multi-modal Large Language Models (MLLMs) in single-image morphing attack detection (S-MAD). The analysis encompasses a diverse set of recent MLLMs and compares them against task-specific state-of-the-art morphing detectors developed under various design paradigms and supervision regimes. Our results reveal that MLLMs, although originally trained for general visual–linguistic reasoning rather than biometric forensics, exhibit a remarkable latent sensitivity to morphing artifacts. This suggests that large-scale multimodal alignment endows these models with transferable forensic reasoning abilities—linking semantic understanding to subtle image inconsistencies without explicit supervision. Notably, LLaVA1.6-Mistral-7B achieves a new state of the art, surpassing both other open-source MLLMs and highly optimized task-specific MAD architectures. Beyond detection accuracy, MLLMs offer several advantages over conventional MAD approaches. They operate in a truly zero-shot manner, adapt to unseen attack types through prompt-based conditioning, and are able to produce human-interpretable outputs for more transparent and verifiable decision making. Such explainable responses are particularly valuable in security-critical or legal forensic contexts where interpretability, accountability and trustworthiness are essential. Furthermore, the ability of a single foundation model to generalize across different biometric manipulation tasks highlights its potential as a unified framework for visual integrity assessment. These findings open a promising direction for the use of foundation models in biometric forensics. Future work will explore lightweight adaptation or fine-tuning strategies to further enhance accuracy while preserving interpretability and reducing computational cost. 

\vspace{-5px}
\section{Declaration of generative AI and AI-assisted technologies in the writing process}
During the preparation of this work the authors used ChatGPT in order to improve the readability and language of the manuscript. After using this tool/service, the authors reviewed and edited the content as needed and take full responsibility for the content of the published article.
\vspace{-15px}
\section*{Funding}
\vspace{-5px}
The research presented in this work was supported in parts by the Slovenian Research Agency (ARIS) Research Programmes P2-0250(B) "Metrology and Biometric Systems", and the ARIS Research Project J2-50065 "DeepFake DAD".
\vspace{-15px}
\bibliographystyle{elsarticle-num-names}  
\bibliography{thebibliography}        
\end{document}